\def\year{2020}\relax
\documentclass[letterpaper]{article} 
\usepackage{aaai20}  
\usepackage{times}  
\usepackage{helvet} 
\usepackage{courier}  
\usepackage[hyphens]{url}  
\usepackage{graphicx} 
\usepackage{graphicx}  
\usepackage{slashbox}
\usepackage{mathtools}
\usepackage{amsmath, amsthm, amssymb, amsfonts, amsopn}
\usepackage{graphicx} 
\usepackage{textcomp}
\usepackage{xfrac}
\usepackage{bbm}
\usepackage{overpic}
\usepackage{subfig}
\usepackage{multirow}
\newcommand{\supl}[1]{{#1}}
\newcommand{\N}{\mathcal{N}}

\newcommand{\Figure}[1]{Figure~\ref{fig:#1}}

\newcommand{\Eq}[1]{Eq.~(\ref{eq:#1})}
\newcommand{\eq}[1]{(\ref{eq:#1})}

\title{NeoNav: Improving the Generalization of Visual Navigation \\ via Generating Next Expected Observations}
\author{Qiaoyun Wu\textsuperscript{\rm 1,2}, Dinesh Manocha\textsuperscript{\rm 2}, Jun Wang\textsuperscript{\rm 1}, Kai Xu\textsuperscript{\rm 3}\thanks{Corresponding author: Kai Xu (kevin.kai.xu@gmail.com)}\\
\textsuperscript{\rm 1}Nanjing University of Aeronautics and Astronautics, \textsuperscript{\rm 2}The University of Maryland,\\ \textsuperscript{\rm 3}National University of Defense Technology\\
(Project page: \url{http://kevinkaixu.net/projects/neonav.html})}

\begin{document}

\maketitle

\begin{abstract}
We propose improving the cross-target and cross-scene generalization of visual navigation
through learning an agent that is guided by conceiving the next observations it expects to see.
This is achieved by learning a variational Bayesian model, called \emph{NeoNav}, which \emph{generates} the \emph{next expected observations (NEO)} conditioned on the current observations of the agent and the target view.  Our generative model is learned through optimizing a variational objective encompassing two key designs. First, the latent distribution is conditioned on current observations and the target view, leading to a model-based, target-driven navigation. Second, the latent space is modeled with a Mixture of Gaussians conditioned on the current observation and the next best action. Our use of mixture-of-posteriors prior effectively alleviates the issue of over-regularized latent space, thus significantly boosting the model generalization for new targets and in novel scenes. Moreover, the NEO generation models the forward dynamics of agent-environment interaction, which improves the quality of approximate inference and hence benefits data efficiency. We have conducted extensive evaluations on both real-world and synthetic benchmarks, and show that our model consistently outperforms the state-of-the-art models in terms of success rate, data efficiency, and generalization.
\end{abstract}


\section{Introduction}
Mapless visual navigation is an important skill for robots operating in unknown, unstructured environments.
It is characterized as the ability of a robot to navigate itself from an arbitrary location in the environment
to a goal position, based solely on the visual inputs from its on-board sensors.
The main challenge of visual navigation lies in understanding the scene layout
based on the visual observations and reasoning about the spatial relation between the current and
the target location. This is the main impediment hindering the generalization of navigation ability
across different scenes due to visual and structural discrepancy.

Recent years have witnessed fast advancement of visual navigation thanks to deep learning,
e.g., deep reinforcement learning (RL) models~\cite{zhang2017,zhu2017,gupta2017,racaniere2017imagination}.
Model-free approaches learn to directly map raw observations to values or actions, which
usually suffers from low data efficiency.
Model-based methods tackle this issue through modeling the transition dynamics of agent-environment interaction.
Such model can be used to reason about the future, thus relieving the trial-and-error learning endeavor.
However, it is difficult to learn a powerful model that generalizes across different scenes,
which is known as the model imperfection issue~\cite{zhu2018scores}.

We propose \emph{NeoNav}, a model-based, supervised learning approach to visual navigation with strong model generality. In our method, the agent is guided by conceiving the next observations it expects to see supposing the best action is taken.
This is realized by learning a \emph{generative model} conditioned on the multi-view observations at the current location as well as the target view, from which the next expected observation (NEO) can be generated.
We predict the next best action based on the generated NEO and the current (front-view) observation.
We frame this problem as a variational Bayesian inference where the variational lower-bound (objective)
consists of three terms: reconstruction, regularization and classification.
The minimization of the reconstruction error maximizes the likelihood of the NEO given the current observations and target view.
The regularization term drives the variational posterior to match a prior distribution.
The classification term is devised for action prediction.

The key characteristic of our approach is the modeling of the latent space.
First, to enable target-driven navigation, the latent distribution is conditioned on current observations and target views.
Second, we model the latent space with a Mixture of Gaussians conditioned on current observations and next best actions. Such a variational \emph{mixture of posteriors} prior~\cite{tomczak2017} effectively alleviates
over-regularization of the latent space, thus facilitating cross-scene model generalization.
Moreover, NEO generation via sampling over the latent space essentially models the \emph{forward dynamics} of the agent-environment interaction, i.e. action-driven state transition.
This improves the expressiveness of the approximation of variational inference~\cite{cremer2018inference}, thereby greatly enhancing inference generalization and data efficiency.

Although our supervised setting requires denser training signals, the learned model shows significantly better generality over unseen test scenes than RL-based approaches (even if enhanced by supervision such as in behavior cloning).
Fortunately, target-driven navigation tasks enjoy the easy acquisition of
ground-truth paths as training data (e.g., using A$^\ast$ algorithm).
We conducted extensive evaluations on public datasets of both synthetic (AI2-THOR framework~\cite{zhu2017}) and real-world (Active Vision Dataset, AVD~\cite{ammirato2017dataset}) scenes. We demonstrate that our model attains at least a $5\%$ higher success rate for both cross-target and cross-scene generalization, compared to several state-of-the-art alternative methods, ranging from model-based to model-free, from RL-based to supervised, and from target-driven to semantic-driven.


\section{Related Works}
\paragraph{Model-free navigation.}
This approach learns to map the raw observations directly to actions.
Mnih et al.~\shortcite{mnih2015} present the first deep reinforcement learning model, called Deep Q-Learning, that successfully learns control policies directly from high-dimensional sensory input.
Schulman et al.~\shortcite{schulman2015} propose the Trust Region Policy Optimization (TRPO), which is effective for optimizing large nonlinear policies and demonstrates robust performance on a wide variety of robotic tasks.
Lillicrap et al.~\shortcite{lillicrap2015} present the DDPG (Deep Deterministic Policy Gradient), which can robustly solve many simulated tasks.
Model-free methods usually require large training data and the policies do not readily generalize to novel tasks in unseen environments.

Several works study using deep neural networks to realize classical iterative planning without an explicit environmental model~\cite{tamar2016,silver2017,oh2017,lee2018,zhao2018triangle}.
Zhang et al.~\shortcite{zhang2017} focus on the problem of robot navigation in maze-like environments and present a successor-feature-based deep RL algorithm that can transfer navigation policies across similar environments. Most existing models are trained and tested on mazes; an exception is~\cite{mirowski2018learning} which proposes a deep RL model for navigating in cities.
Zhu et al.~\shortcite{zhu2017} propose an excellent feed-forward architecture for target-driven visual navigation by combining a Siamese network with the A3C algorithm~\cite{mnih2016}. They focus on cross-target generalization in smaller indoor scenes and do not consider generalization to previously unseen environments.
In~\cite{mousavian2018}, semantic scene segmentation is incorporated in learning to map from semantic information to navigation actions. Through comparison, we show that our method has better cross-target and cross-scene generalization.

\begin{figure*}
\begin{center}
\includegraphics[width=0.95\textwidth]{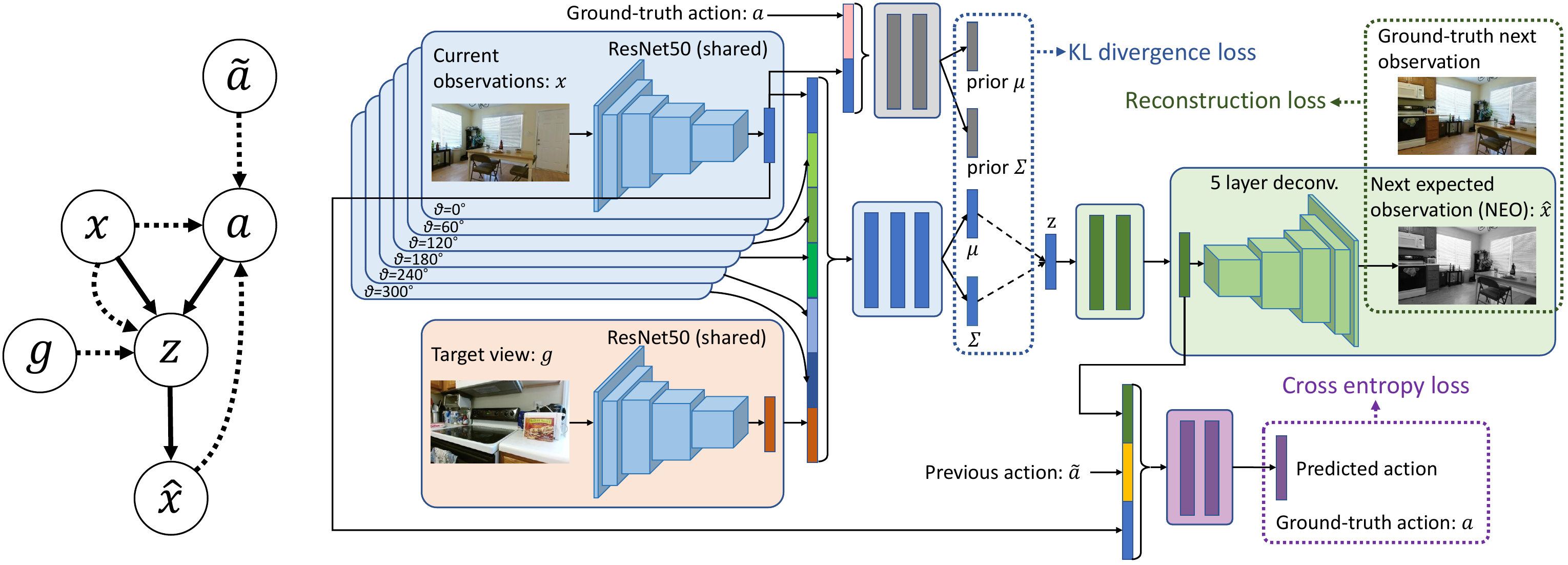}\vspace{-12pt}
\end{center}
   \caption{Model overview: the probabilistic graphical model and the network architecture. In the graphical model, the posterior $p_{\theta}(z|x,a)$ and generator $p_{\theta}(\hat{x}|z)$ are denoted with solid lines, while the variational approximation $q_{\lambda}(z|x,g)$ and the action prediction $q_{\varphi}(a|x,\hat{x},\tilde{a})$ are depicted with dashed lines. The generative model is realized with a variational auto-encoder architecture: The encoder takes the current observations $x$ and the target view $g$ as input. The decoder generates the NEO $\hat{x}$ from a random vector sampled from the latent space defined by the Gaussian $\N(\mu, \Sigma)$. The feature in the decoding module is used in predicting the next action. Three losses used for learning the generative model are marked with dashed boxes.}
\label{fig:overview}\vspace{-12pt}
\end{figure*}

\paragraph{Model-based navigation.}
This approach achieves better data efficiency, but has the issues of cross-scene generalization due to
model imperfections. Several approaches have been proposed to address the model imperfection issue,
such as capturing model uncertainty~\cite{deisenroth2011pilco,marco2017virtual} and incorporating semantic priors into environmental models~\cite{yang2018,mousavian2018}.
With the advances of attention mechanisms in deep learning, many works propose
modeling the environment with a memory unit.

Savinov et al.~\shortcite{savinov2018} introduce a topological landmark-based memory for navigation.
A common issue with such an approach is that the memory, representing an allocentric map of the scene,
grows in size as the scene exploration proceeds, limiting its practical utility in navigating within large environments.
In~\cite{gupta2017}, the problem is alleviated by learning an ego-centric mapper and planner,
which, however, assumes perfect odometry.
Henriques and Vedaldi~\shortcite{henriques2018mapnet} develop a differentiable
module that is able to associate an egocentric representation of a scene to an allocentric one.
%
Our method models the environment with the probabilistic latent distribution in a variational Bayesian framework, where both model generality and model scalability are attained by imposing a mixture-of-posteriors prior.
A similar model was proposed in~\cite{henriques2018mapnet} in the RL setting.

\paragraph{Imagination-based navigation.}
Some model-based navigation methods reason about the future based on the internal model.
Razvan et al.~\shortcite{pascanu2017learning} introduce an imagination-based planner, which  is the first model-based, sequential decision-making agent that can learn to propose (imagine), evaluate, and execute plans.
The method demonstrates good performance on 2D maze-solving tasks.
Imagination-Augmented Agents (I2As)~\cite{racaniere2017imagination} was later proposed and it learns to generate and interpret predictions as additional context for deep policy networks.
These methods are generally data-efficient, but have difficulty in scaling to complex, high-dimensional tasks.
Generally, similar ideas have been well practiced in the studies of deep RL, where
the internal model is used to predict future observations and/or rewards~\cite{oh2015action,leibfried2016,dosovitskiy2016,jaderberg2016,mirowski2016,finn2017,pathak2017curiosity}.
Watter et al.~\shortcite{watter2015} introduce Embed to Control (E2C), which learns to generate image trajectories from a latent space in which the dynamics is constrained to be locally linear, in contrast to the non-linear dynamics modeled by our latent space. Although sharing a similar spirit, our work is significantly different from the imagination-augmented RL-based navigation~\cite{racaniere2017imagination,pascanu2017learning}. First, our model is formulated as a variational Bayesian inference trained with supervised learning rather than RL. Second, their imaginations refer to a simulated rollout of trajectories, while our imagination is a one-step imagination of the next observation.


\section{Method}

\subsection{Problem Setting}
Target-driven visual navigation takes the current observations $x$ captured by the agent
and a target view $g$ as input,
and predicts the next best action $a$ at each time step to navigate the agent, until reaching
the target position.

\paragraph{Observations and goals.}
The agent camera has only the azimuth DoF.
At each agent location, the observation $x$ consists of $K$ views with evenly distributed azimuth angles: $\{0^\circ, \frac{1}{K}360^\circ, \ldots, \frac{K-1}{K}360^\circ\}$, in which $0^\circ$ corresponds to the front-looking view.
The agent captures an image (RGB, depth or RGB-D) at each view.
The $K$-view observations provide a local context of the environment, based on which
the agent is able to reason about its location and the room layout of its surroundings.
The target view is consistent with the observation views in terms of image data modality.

\paragraph{Action space.}
At each time step, the agent can choose one action from a discrete set of allowable actions: $\{move\_forward, move\_back, move\_left, move\_right,
\\ rotate\_ccw, rotate\_cw, stop\}$,
where $move$ means horizontal movement of the agent and $rotate$ refers to azimuth rotation of the camera.
$ccw$ and $cw$ stands for counter-clockwise and clockwise, respectively.
If there is no next observation view associated with an action, the action is considered to cause a collision.

\subsection{The Variational Bayesian Navigation Model}
Given the current observation $x$, instead of directly predicting the next best action $a$ as in many other works,
we opt to first \emph{generate} the next expected observation (NEO) $\hat{x}$ assuming that the next best action $a$ is known \emph{a priori} and is executed. This can be described with a generative model:
\begin{equation}\label{eq:generative}
p_{\theta}(\hat{x}, z|x,a)=p_{\theta}(\hat{x}|z)p_{\theta}(z|x,a),
\end{equation}
where $p_{\theta}(\hat{x}, z|x,a)$ is a parametric model of the joint distribution
over the NEO $\hat{x}$ and a latent variable $z$.
Essentially, this generative model is a probabilistic \emph{forward dynamics model} of the agent,
where the acquirement of the next observation is driven by the selected next action.

To learn the generative model, one typically maximizes the marginal log-likelihood $\log{p_{\theta}(\hat{x}|x,a)}$. However, when the model is parameterized by a
neural network, the optimization could be difficult due to the intractability of the marginal likelihood.
Moreover, the next best action $a$ is unknown \emph{a priori} and is inherently determined by the target $g$.
To this end, we apply \emph{variational inference} and
introduce an inference network $q_{\lambda}(z|x,g)$ with parameters $\lambda$ to approximate the true posterior $p_{\theta}(z|x,a)$.
In particular, we optimize the following lower bound of the marginal likelihood:
\begin{equation}\label{eq:lower_bound}
\log{p_{\theta}(\hat{x}|x,a)}\geq \mathbb{E}_{z\thicksim q_{\lambda}(z|x,g)}[\log{\frac{p_{\theta}(\hat{x},z|x,a)}{q_{\lambda}(z|x,g)}}] = \mathcal{L}(\hat{x}).
\end{equation}
This lower bound forms our objective function:
\begin{equation}\label{eq:obj_func}
\begin{aligned}
\mathcal{J}= &- \mathbb{E}_{z\thicksim q_{\lambda}(z|x,g)}[\log{p_{\theta}(\hat{x}|z)}] \\
&+ KL[q_{\lambda}(z|x,g)||p_{\theta}(z|x,a)] = -\mathcal{L}(\hat{x}),
\end{aligned}
\end{equation}
where $KL$ denotes the Kullback-Leibler divergence.
During training, $p_{\theta}(z|x,a)$ can be estimated as a Gaussian distribution conditioned on the current observation $x$ and the ground-truth action $a$, leading to a \emph{mixture-of-posteriors prior} imposed on the latent distribution.

To realize robot navigation, we learn a navigation action classifier
$q_{\varphi}(a|x,\hat{x},\tilde{a})$ which predicts the next best action $a$ based on
the current observation $x$ , the generated NEO $\hat{x}$ as well as the previous action $\tilde{a}$.
Integrating action prediction, the objective function becomes:
\begin{equation}\label{eq:obj_func_action}
\begin{aligned}
\mathcal{J} = &- \alpha \mathbb{E}_{z\thicksim q_{\lambda}(z|x,g)}[\log{p_{\theta}(\hat{x}|z)}] \\
&+ \beta KL[q_{\lambda}(z|x,g)||p_{\theta}(z|x,a)] \\
&+ \gamma \mathbb{E}_{a\thicksim p(a)}[-\log{q_{\varphi}(a|x,\hat{x},\tilde{a})}],
\end{aligned}
\end{equation}
where $a\thicksim Cat(1/C)$. \supl{A complete derivation of this objective is given in the supplemental material.}
The objective function in \eq{obj_func_action} is composed of a reconstruction loss, a KL divergence loss and a cross entropy loss. The three hyper-parameters are empirically set as $\alpha=0.01$, $\beta=0.0001$ and $\gamma=1$ throughout our experiments.
\Figure{overview}(left) shows the probabilistic graphical model of our navigation model.

\subsection{The Network Architecture}
Corresponding to the variational objective, the architecture of our network consists of three subnetworks (see \Figure{overview}).
The \emph{variational inference module} takes the full observation views at the current robot position
as well as the target view as input and extracts a $2048$-D feature vector for each of them using a ResNet-50.
The input image resolution is $64*64$. These output $2048$-D feature vectors are then used to infer a vector of latent variables of dimension $400$~with a MLP. Here, a KL divergence loss is minimized
to impose the distribution of the latent variables to match a prior estimated from
the current observation (front view only) and the ground-truth action.
The \emph{NEO generation module} then generates the NEO in the front view out of a latent vector,
using a two-layer MLP followed by a 5-layer convolutional network (\supl{please refer to the supplemental material for details}).
This task is trained with the supervision of ground-truth next observation.
The \emph{action prediction module} maps the concatenation of the last layer feature of the NEO generation module ($2048$-D), the feature of the current observation ($2048$-D) and the feature ($1024$-D) extracted from the previous action ($7$-D one-hot vector) into the predicted next action ($7$-D), using a four-layer MLP.
Ground-truth actions are used to train this subnetwork.

\paragraph{Model training and testing.}
Our model is trained and tested with both real-world environments from the Active Vision Dataset (AVD)~\cite{ammirato2017dataset} and synthetic scenes of AI2-THOR~\cite{zhu2017}. Each scene in the dataset is represented as a grid of robot locations (see \Figure{path}).
The size of the grid cell is $0.25$-$0.5$ meters. For each grid point, $6$ azimuth camera views are captured for AVD and $4$ for AI2-THOR.
For the task of target-driven navigation, the ground-truth navigation path is simply the shortest path over the grid. The optimization of the variational objective is achieved by Monte Carlo sampling, where the gradients are backpropagated with the standard reparameterization trick~\cite{kingma2013}.


At test time, our model is used as a controller for the agent to predict the next action given the current observations. We feed the current observation views and the target view into the inference module to obtain
a Gaussian component in the latent space. The features used for NEO prediction and extracted for the current front-view observation, as well as the previous action, are used for next action prediction.
The actual generation of NEO, however, is not needed in testing.


\section{Experiments}
We evaluate both cross-target and cross-scene generalization,
as well as a few other important characteristics, of our model by comparing it with
one baseline and a few state-of-the-art methods.
We also compare our method to two ablated variants of it to justify our major design choices.
In addition, we visualize the latent space of our model for a better understanding of what we learn,
as well as the navigation paths for a qualitative evaluation.

\paragraph{Experimental settings.}
Our evaluations are conducted on both AVD and AI2-THOR.
AVD contains $11$ relatively complex real-world houses, of which
$8$ houses were used for training and $3$ for testing.
AI2-THOR contains $120$ scenes in four categories including
kitchen, living room, bedroom, and bathroom.
Each category includes $30$ scenes, out of which $20$ are used for training and $10$ for testing.
For all the methods being compared, we train a single model for all the AVD scenes and separate models for the categories of AI2-THOR.

For each training scene, we choose fifteen different views as the target,
each of which contains a targeted object such as a dining table, a refrigerator, a sofa, a television, etc.
During testing, the target views are randomly sampled from the test scenes, encompassing both views similar to trained targets and views unseen in training.
When sampling start points, we consider the ratio of the shortest path distance to the Euclidean distance between start and goal positions~\cite{savva2019habitat}.
We perform aggressive rejection sampling to ensure that $15\%$ of the tasks have a ratio within the range of $[1, 1.1]$.
As in many navigation systems, a collision detection module is devised.
When a collision is detected, the action with the next largest probability is chosen.

\paragraph{Success criteria.}
In each episode, the agent runs until arriving at the goal (the distance to the target position
is less than $1$ meter and the angle between the current and target view direction is less than $90^\circ$),
reaching the maximum number of steps ($100$), or issuing a $stop$ action.
In the setting with $stop$ action, an episode is successful if and only if
the agent issues a $stop$ action exactly when it reaches the goal.
This success criterion is apparently stricter than that in the setting without a $stop$ action.
We will evaluate both cases.

\paragraph{Evaluation metrics.}
We adopt two evaluation metrics,
success rate and success weighted by (normalized inverse) path length (SPL)~\cite{anderson2018}.
Success rate is the fraction of the runs that successfully navigate to the goal.
SPL is defined as $\frac{1}{N}\sum_{i=1}^{N}S_i\frac{L_i}{\max\{P_i,L_i\}}$,
where $N$ is the number of navigation tasks, $S_i$ a binary indicator of success in the $i$-th task.
$P_i$ and $L_i$ denote the actual path length and the shortest path distance for the $i$-th task, respectively.


\begin{table*}
\centering
\caption{Navigation performance (success rate and SPL, in $\%$) comparison on \emph{novel scenes} from AVD with $stop$ action.}\vspace{-6pt}
\label{tab:diffscene}
\resizebox{0.95\textwidth}{!}{
\begin{tabular}{l|c|c|c|c|c||c}
\cline{1-6}
\hline
\backslashbox{Model}{Target} & Table & Exit & Couch & Refrigerator  & Sink & Avg. \\\cline{1-7}
\hline
\hline
Random Walk &  4.0 / 2.7 &  4.6 / 3.1&  3.4 / 2.1&  3.2 / 2.1&  3.6 / 2.7 &  3.8 / 2.7\\\hline
TD-A3C-U&  5.3$\pm$0.6 / 2.4$\pm$0.3 &  6.7$\pm$0.4 / 4.2$\pm$0.2&  4.3$\pm$0.4 / 2.5$\pm$0.2 &  5.6$\pm$0.3 / 2.7$\pm$0.2 &  7.1$\pm$0.5 / \textbf{3.6}$\pm$0.2&  5.8 / 3.1 \\\hline
TD-A3C&  \textbf{12.4}$\pm$2.1 / 1.5$\pm$0.6 &  23.0$\pm$1.6 / 2.8$\pm$0.5&  \textbf{15.0}$\pm$1.9 / 1.7$\pm$0.2 &  7.2$\pm$1.1 / 1.1$\pm$0.2 &  \textbf{13.4}$\pm$1.4 / 1.7$\pm$0.3&  14.2 / 1.8 \\\hline
Gated-LSTM-A3C&  6.5$\pm$0.7 / 1.3$\pm$0.3 &  16.2$\pm$0.8 / 4.0$\pm$0.2&  8.0$\pm$0.5 / 1.2$\pm$0.3&  14.3$\pm$0.7 / 3.3$\pm$0.4 & 6.5$\pm$0.5 / 0.9$\pm$0.2&  10.3 / 2.1\\\cline{1-7}
I2A&  10.6$\pm$1.1 / 1.3$\pm$0.3 &  21.7$\pm$1.3 / 2.2$\pm$0.4&  14.3$\pm$0.9 / 2.1$\pm$0.3 &  8.9$\pm$0.7 / 2.3$\pm$0.3 & 11.2$\pm$0.9 / 1.6$\pm$0.3&  13.3 / 1.9\\\hline
Ours&  12.1$\pm$1.3 / \textbf{3.4}$\pm$0.8 & \textbf{30.4}$\pm$1.6 / \textbf{8.4}$\pm$0.9 & 11.0$\pm$1.1 / \textbf{2.9}$\pm$0.9 & \textbf{35.0}$\pm$1.4 / \textbf{11.9}$\pm$1.0 & 11.0$\pm$0.9 / 2.7$\pm$0.4 &  \textbf{19.9} / \textbf{5.9}\\\hline
\end{tabular}}\vspace{-6pt}
\end{table*}

\begin{table*}
\centering
\caption{Navigation performance (success rate and SPL, in $\%$) comparison on \emph{novel scenes} from AVD without $stop$ action.}\vspace{-6pt}
\label{tab:simplercase}
\resizebox{.95\textwidth}{!}{
\begin{tabular}{l|c|c|c|c|c||c}
\cline{1-6}
\hline
\backslashbox{Model}{Target} & Table & Exit & Couch & Refrigerator  & Sink & Avg. \\\cline{1-7}
\hline
\hline
Random Walk & 34.8 / 12.9 &  29.0 / 11.3&  29.8 / 10.8&  27.4 / 10.7&  23.0 / 10.2 &  28.8 / 11.2\\\hline
TD-A3C-U&  39.7$\pm$4.4 / 13.1$\pm$2.6 &  29.3$\pm$4.3 / 10.9$\pm$2.1&  30.1$\pm$4.1 / 9.9$\pm$2.0 &  28.4$\pm$3.3 / 10.1$\pm$1.7 &  22.9$\pm$3.0 / 9.7$\pm$1.9&  30.1 / 10.7 \\\hline
TD-A3C&  45.8$\pm$5.3 / 5.8$\pm$1.7 &  37.6$\pm$3.6 / 6.3$\pm$1.4 &  37.2$\pm$4.0 / 5.0$\pm$1.3 &  16.8$\pm$3.2 / 4.4$\pm$1.0 &  23.4$\pm$3.7 / 4.7$\pm$1.9&  32.2 / 5.2 \\\hline
Gated-LSTM-A3C&  31.0$\pm$3.3 / 8.6$\pm$1.0 &  31.1$\pm$2.7 / 13.7$\pm$1.5&  25.3$\pm$2.3 / 5.8$\pm$1.1&  31.4$\pm$2.9 / 12.9$\pm$1.9& 23.0$\pm$2.4 / 8.3$\pm$1.0&  28.4 / 9.9\\\hline
I2A&  40.6$\pm$4.2 / 6.7$\pm$2.1 &  37.3$\pm$3.6 / 6.9$\pm$2.0&  35.9$\pm$3.3 / 6.3$\pm$1.7 &  17.3$\pm$3.1 / 5.6$\pm$1.2 & 20.1$\pm$3.0 / 6.3$\pm$1.4&  30.2 / 6.4\\\hline
Ours&  57.6$\pm$4.5 / 32.6$\pm$2.9 & \textbf{52.4}$\pm$4.1 / \textbf{23.9}$\pm$2.7 & \textbf{43.4}$\pm$3.9 / \textbf{21.1}$\pm$2.1 & \textbf{46.4}$\pm$3.4 / \textbf{24.3}$\pm$1.9& \textbf{38.8}$\pm$3.3 / \textbf{25.8}$\pm$1.7 &  \textbf{47.7} / \textbf{25.5}\\\hline
\hline
Ours-FrontView& 49.9$\pm$4.1 / 24.3$\pm$1.6  & 40.6$\pm$3.9 / 13.4$\pm$ 1.2 & 38.1$\pm$4.0 / 16.1$\pm$ 1.5 & 28.7$\pm$3.4 / 12.1$\pm$1.3  & 28.1$\pm$3.5 / 11.9$\pm$ 2.1 &37.1 / 15.6\\\hline
Ours-NoGen&  43.6$\pm$2.1 / 28.6$\pm$0.9 &  34.3$\pm$2.3 / 19.4$\pm$0.7&  31.7$\pm$1.9 / 18.4$\pm$0.7 &  37.7$\pm$2.0 / 20.1$\pm$1.1 &  26.8$\pm$2.6 / 22.0$\pm$1.5&  34.8 / 21.7  \\\hline
Ours-NoMoP&  50.7$\pm$5.3 / 28.2$\pm$3.6 &  49.9$\pm$4.2 / 22.8$\pm$3.1&  42.9$\pm$4.3 / 20.4$\pm$2.9 &  34.4$\pm$3.9 / 19.9$\pm$2.0 &  23.7$\pm$3.7 / 17.2$\pm$2.1 &  40.3 / 21.7 \\\hline
\hline
Ours (RGB)&  48.4$\pm$4.9 / 25.7$\pm$3.1 &  32.4$\pm$3.1 / 16.7$\pm$2.0 &  28.7$\pm$2.7 / 16.8$\pm$1.4 &  37.9$\pm$4.5 / 18.8$\pm$2.9 &   38.2$\pm$2.3 / 26.4$\pm$1.1 &  37.1 / 20.9 \\\hline
Ours (RGBD)& \textbf{59.6}$\pm$4.3 / \textbf{36.8}$\pm$3.0&  38.4$\pm$4.4 / 18.5$\pm$2.9&  31.5$\pm$3.7 / 17.8$\pm$1.6&  43.5 $\pm$4.1 / 15.3$\pm$2.4&  37.8$\pm$3.2 / 21.7$\pm$2.0&  42.2 / 22.0 \\\hline
\end{tabular}}\vspace{-6pt}
\end{table*}

We compare with the following baselines/alternatives:
\begin{itemize}
  \item \textbf{Random Walk}, a baseline where the agent randomly chooses an action at each time step.
  \item \textbf{TD-A3C}, a target-driven visual navigation model based on deep RL~\cite{zhu2017}. The reactive policy is trained with the views of the previous three steps in addition to the current view. In addition, we did not freeze the ResNet-50 when training the model. We compare against two variants of the model, with or without supervision, denoted as TD-A3C and TD-A3C-U, respectively.
  \item \textbf{I2A}, i.e., Imagination-Augmented Agents~\cite{racaniere2017imagination}, which is a model-based deep RL model. The original method is developed for 2D maze-solving tasks; we re-implemented it for visual navigation by changing its input to first-person views.
  \item \textbf{Gated-LSTM-A3C}, an LSTM-based variant of A3C model adapted from~\cite{wu2018}, where we train the model with back-propagation through time over $10$ unrolled time steps; the goal is specified as an image.
  \item \textbf{TD-Semantic}, a state-of-the-art target-driven navigation model based on deep supervised learning. The method leverages the semantic and contextual representations obtained by off-the-shelf object detection and segmentation methods~\cite{mousavian2018}.
  \item \textbf{Ours-FrontView}, a variant of ours where the current observation at each time step is only the front view rather than four views.
  \item \textbf{Ours-NoGen}, a non-generative variant of our model where the next expected observation is predicted directly from the current observations and the target view. This is implemented simply by removing the Gaussian sampling and the KL-loss in \Eq{obj_func_action}.
  \item \textbf{Ours-NoMoP}, a baseline variant of our model in which the latent space follows the standard normal distribution prior, instead of the mixture-of-posteriors prior.
\end{itemize}
Except for the Random Walk and TD-A3C-U, all alternatives are trained with supervision.
TD-A3C, I2A and Gated-LSTM-A3C are all first trained via behavioral cloning using ground-truth paths.
After pre-training, we update the three policy layers using a shaped reward based on the geodesic distance to the goal, $geo(x,g)$, as described in~\cite{gordon2019}:
$
r_t=geo(x_{t-1},g)-geo(x_t,g)+\zeta,
$where $\zeta=-0.01$ is a small constant time penalty.
\supl{More implementation details are provided in the supplemental material.
Unless explicitly stated otherwise, all methods take depth images as the input observation. For our model, we also implement variants taking RGB and/or RGBD images as input.}

\begin{figure}
\begin{center}
\includegraphics[width=.95\columnwidth]{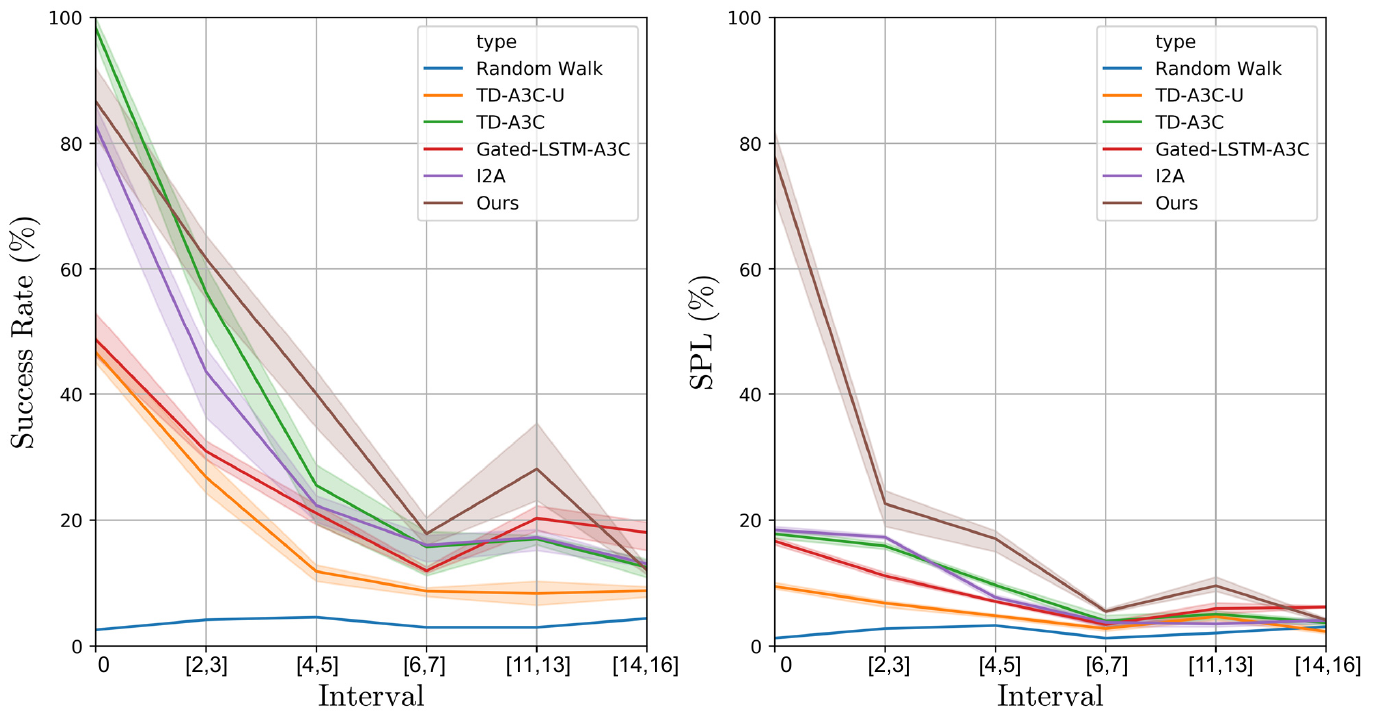}
\end{center}\vspace{-6pt}
   \caption{Navigation performance (success rate and SPL, in $\%$) comparison for novel targets on AVD with $stop$ action.
 }\vspace{-12pt}
\label{fig:noveltarget}
\end{figure}

\paragraph{Cross-target generalization.}
Over the $8$ training scenes of AVD, we evaluate navigation performance
for $40$ novel targets that are unseen in the training phase.
These targets are classified into five intervals of the shortest distance between the test and the nearest trained targets: $[2,3]$, $[4,5]$, $[6,7]$, $[11,13]$, and $[14,16]$.
For each interval, we sample $1000$ navigation tasks with different starting points.
The results on the two metrics (with standard deviation measured from five times training) are reported in Figure~\ref{fig:noveltarget}.
Generally, the success rate decreases as the distance between the test and trained targets increases.
Our model with default depth input outperforms the state-of-the-art alternatives by $>5\%$ for average success rate and by $>4\%$ for average SPL.
An observation is that the success rate is related to the degree of presence of the targeted object in the target views.
If the targeted object is completely visible in the target view, the target is more instructive and thus easier to reach.

\paragraph{Cross-scene generalization.}
To evaluate model generality over unseen scenes,
we perform navigation with $15$ sampled target views from the testing split of AVD.
The targets are classified into five groups according to the object of interest in the target views; see Table~\ref{tab:diffscene}.
Note that object labels were not used for navigation.
For each group, we sample $1000$ navigation tasks (starting points).
Our model achieves $> 5\%$ higher average success rate and $>2\%$ higher average SPL than the alternative methods.
The standard deviations in the table are measured from training each model for five times.
In Table~\ref{tab:simplercase}, we report the results for the case without a $stop$ action.
\supl{The results of the tasks reversing the start and target points can be found in the supplemental material.}
The plot in Figure~\ref{fig:nav}(left) compares average success rate (without the $stop$ action) of different models over an increasing number of time steps, tested on AVD.
Our method achieves the steepest increase.

Table~\ref{tab:syndiffscene} evaluates target-driven navigation over synthetic scenes from AI2-THOR.
For each of the four room categories, $1000$ randomly generated navigation tasks are sampled from the testing split of the dataset. All methods being compared take RGB input, following the original work~\cite{zhu2017}.
The random walk baseline can be used as a reference to assess the difficulty of the navigation tasks.
For example, living rooms are more challenging while small bathrooms are relatively easy.
For the bathrooms, however, our method fails to beat the TD-A3C, because the transparent glass and texture-less furniture make it difficult for our model to infer the surrounding layout which is important to NEO imagination and action prediction.
Overall, our model has much better cross-scene generality.

\begin{table}[t]
\centering
\caption{Comparing navigation performance (success rate and SPL) on novel scenes from AI2-THOR with $stop$ action.
Note that all methods in this comparison use RGB input.}
\label{tab:syndiffscene}
\resizebox{\columnwidth}{!}{
\begin{tabular}{l|c|c|c|c||c}
\cline{1-6}
\hline
Category & Kitchen & Living room & Bedroom  & Bathroom & Avg. \\\cline{1-6}
\hline
\hline
Random Walk &  7.0 / 3.5 &  1.8 / 1.0&  2.6 / 1.5&  17.9 / 8.0 & 7.3 / 3.5 \\\hline
TD-A3C-U &  8.2 / 3.0 &  2.0 / 1.1 &  3.2 / 1.9 &  19.0 / 9.1&  8.1 / 3.0 \\\hline
TD-A3C &  11.4 / 1.6 &  5.6 / 0.4 &  5.3 / 0.7 &  \textbf{24.3} / 2.3&  11.7 / 1.3 \\\hline
Gated-LSTM-A3C &  13.1 / 3.2 &  4.9 / 1.1 &  5.1 / 1.2 &  19.3 / 7.9&  10.6 / 3.4 \\\hline
I2A &  12.3 / 1.9 &  5.4 / 0.3 &  6.2 / 0.8 &  22.3 / 2.1&  11.5 / 1.3 \\\hline
Ours &  \textbf{19.8} / \textbf{10.6} &  \textbf{11.5} / \textbf{5.3}&  \textbf{13.6} / \textbf{5.9}&  21.9 / \textbf{9.6} &  \textbf{16.7} / \textbf{7.9 }\\
\hline
\end{tabular}}\vspace{-6pt}
\end{table}

\begin{table}[t]
\centering
\caption{Performance (success rate and SPL) for different number of training scenes from AVD without $stop$ action.}
\label{tab:diffscale}
\resizebox{\columnwidth}{!}{
\begin{tabular}{l|c|c|c|c}
\cline{1-5}
\hline
\# Scenes& $8$ & $6$ & $4$ & $2$\\\cline{1-5}
\# Samples& $616,630$ & $524,934$ & $313,652$ & $152,836$\\\cline{1-5}
\hline
\hline
TD-A3C-U&  26.0 / 9.5  & 25.9 / 8.3 &   25.9 / 8.5 &  23.4/ 6.9\\\hline
TD-A3C&  33.4 / 6.1  & 32.1 / 5.9 &   28.4 / 3.9 &  25.9 / 2.8\\\hline
Gated-LSTM-A3C &  25.1 / 9.4  &  24.7 / 8.9 &   20.4 / 6.0& 19.3 / 4.8\\\hline
I2A &  31.4 / 7.1  &  29.7 / 6.9 &   28.2 / 3.2& 26.1 / 2.5\\\hline
Ours & \textbf{47.9} / \textbf{25.8} &\textbf{47.1} / \textbf{24.9} &  \textbf{45.3} / \textbf{22.7}& \textbf{35.1} / \textbf{16.9}  \\\hline
\end{tabular}}\vspace{-6pt}
\end{table}

\paragraph{Ablation study.}
The lower part of Table~\ref{tab:simplercase} shows an ablation study.
Comparing to the front-view only input, four-view input leads to better results.
Our generative method performs much better than its non-generative variant (NoGen)
under the same amount of training.
This conforms to the consensus that learning a stochastic state space is
often more data-efficient than learning a deterministic one~\cite{buesing2018learning}.
The comparison to NoMoP shows that our carefully designed mixture-of-posteriors prior
leads to a more powerful internal model by overcoming the over-regularization of latent space
caused by the commonly used standard normal distribution prior.

\begin{table}[t]
\centering
\caption{Performance (success rate) comparison of semantic-driven navigation on the AVD test split.}
\label{tab:semg}
\resizebox{\columnwidth}{!}{
\begin{tabular}{l|c|c|c|c|c||c}
\cline{1-7}
\hline
Target label & Couch & Table & Refrigerator  & Microwave &TV & Avg. \\\cline{1-7}
\hline
\hline
TD-Semantic (Object)&  80.0 &  38.0&  68.0&  38.0&  44.0 &53.6\\\cline{1-7}
Ours (RGB)& 64.7 &  73.7&  61.3&  38.7&  31.3 & 53.9\\\cline{1-7}
Ours (Depth)& \textbf{83.4} &  67.4&  57.8&  41.1&  \textbf{82.0} & \textbf{66.3}\\\cline{1-7}
Ours (RGBD)& 73.5 &  \textbf{83.5}&  \textbf{72.1}&  \textbf{41.5}&  11.8 & 56.4\\
\hline
\end{tabular}}
\end{table}

\begin{figure}
\begin{center}
\includegraphics[width=\columnwidth]{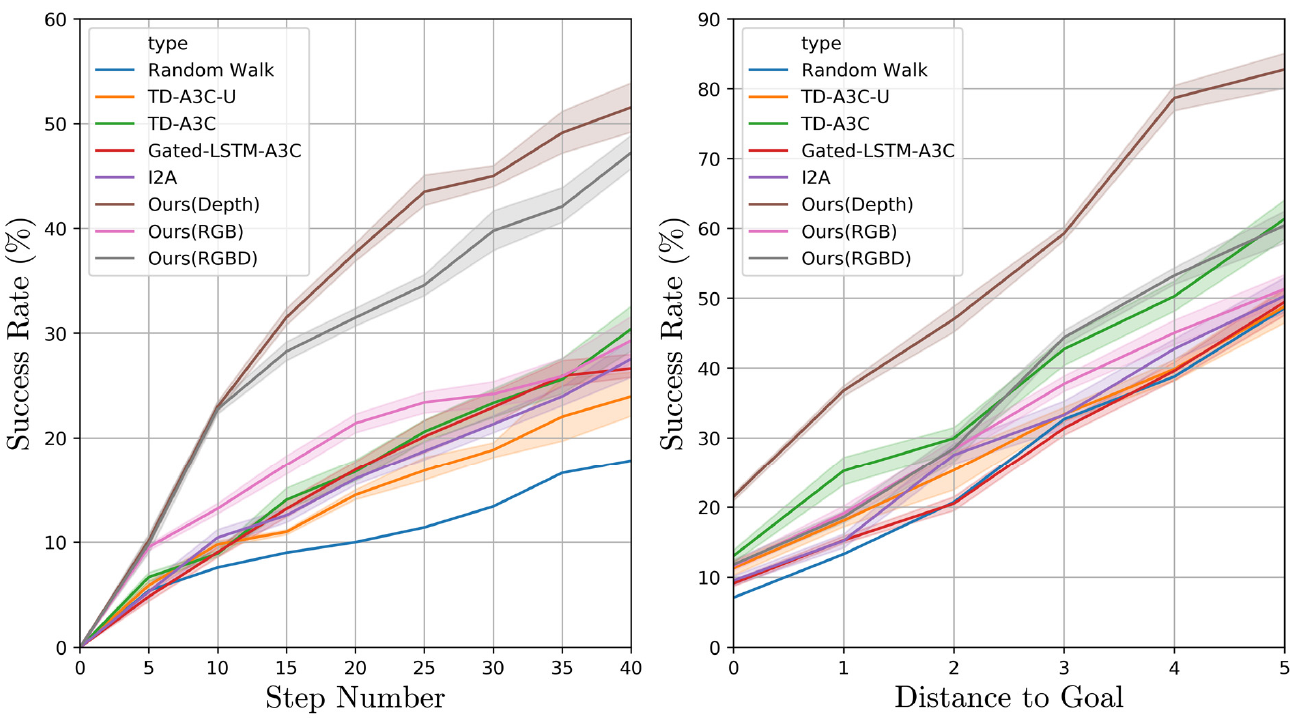}
\end{center}\vspace{-6pt}
   \caption{Left: Success rate over increasing number of time steps. Right: Success rate over different values of distance-to-goal thresholds. Each curve is measured based on $1000$ navigation tasks from the AVD test split.
 }
\label{fig:nav}\vspace{-12pt}
\end{figure}

\paragraph{Input modality.}
Through comparison on different input modalities (Tables~\ref{tab:simplercase}),
the conclusion is that depth information is apparently more useful to our model.
This is because depth images contain rich geometry information which benefits a powerful reasoning about the surrounding layout and the modeling of action-observation dynamics.

\paragraph{Data-efficiency.}
We also evaluate how well our model generalizes when trained on decreasing numbers of scenes (or training samples) from the training split of AVD; see Table~\ref{tab:diffscale}.
The evaluation involves $1000$ different navigation tasks sampled from the testing split of AVD.
All models show increasing success rates and SPLs with increasing numbers of training scenes.
From the results, our method performs consistently better than the alternatives, demonstrating better data-efficiency.
\supl{In the supplemental material, we compare the training curves of the methods.}

\vspace{-6pt}
\paragraph{Close-to-goal stability.}
In most navigation methods, the agent's path tends to oscillate when the agent gets close to the goal.
The main reason is that situations in which the agent is close to the goal are generally sparse in training.
This leads to imbalanced positive and negative situations in training data. Therefore, it is difficult for the trained agent to make a $stop$ decision precisely and decisively when approaching the goal.
Using $1000$ sampled navigation tasks, Figure~\ref{fig:nav}(right) studies the percentage of tasks that succeeds within $40$ time steps (disabling the $stop$ action) over varying distance-to-goal thresholds used for judging navigation success.
In general, smaller thresholds lead to lower success rates due to a higher chance of close-to-goal oscillation.
The plot shows that our method achieves more stable close-to-goal convergence for all thresholds, thanks to the expressive approximation of variational inference learned through modeling the agent-environment interaction.
The latter leads to high data efficiency even for sparse training samples.

\vspace{-6pt}
\paragraph{Navigation driven by semantic labels.}
In methods like TD-Semantic~\cite{mousavian2018}, the navigation goal is defined in the form of a one-hot vector over a prescribed set of semantic labels; for example, $\{Couch, Table, Refrigerator, Microwave, TV\}$.
To compare with TD-Semantic, we adapt our method to take the same navigation goal.
The comparison is conducted on AVD with the same training/testing split,
and the success criterion is within $5$ steps to the goal, as in~\cite{mousavian2018}.
TD-Semantic can learn visual representations for navigation either from RGB and/or depth input or from semantic input of object detection and segmentation obtained by off-the-shelf state-of-the-art methods.
Under the same input modality, our method outperforms TD-Semantic by $23\%$ for RGB input,
$35\%$ for depth input, and $28\%$ for RGBD input for average success rate.
Our best performance (with depth input) is $12.7\%$ higher than theirs with semantic input.

\begin{figure}
\begin{center}
\includegraphics[width=\columnwidth]{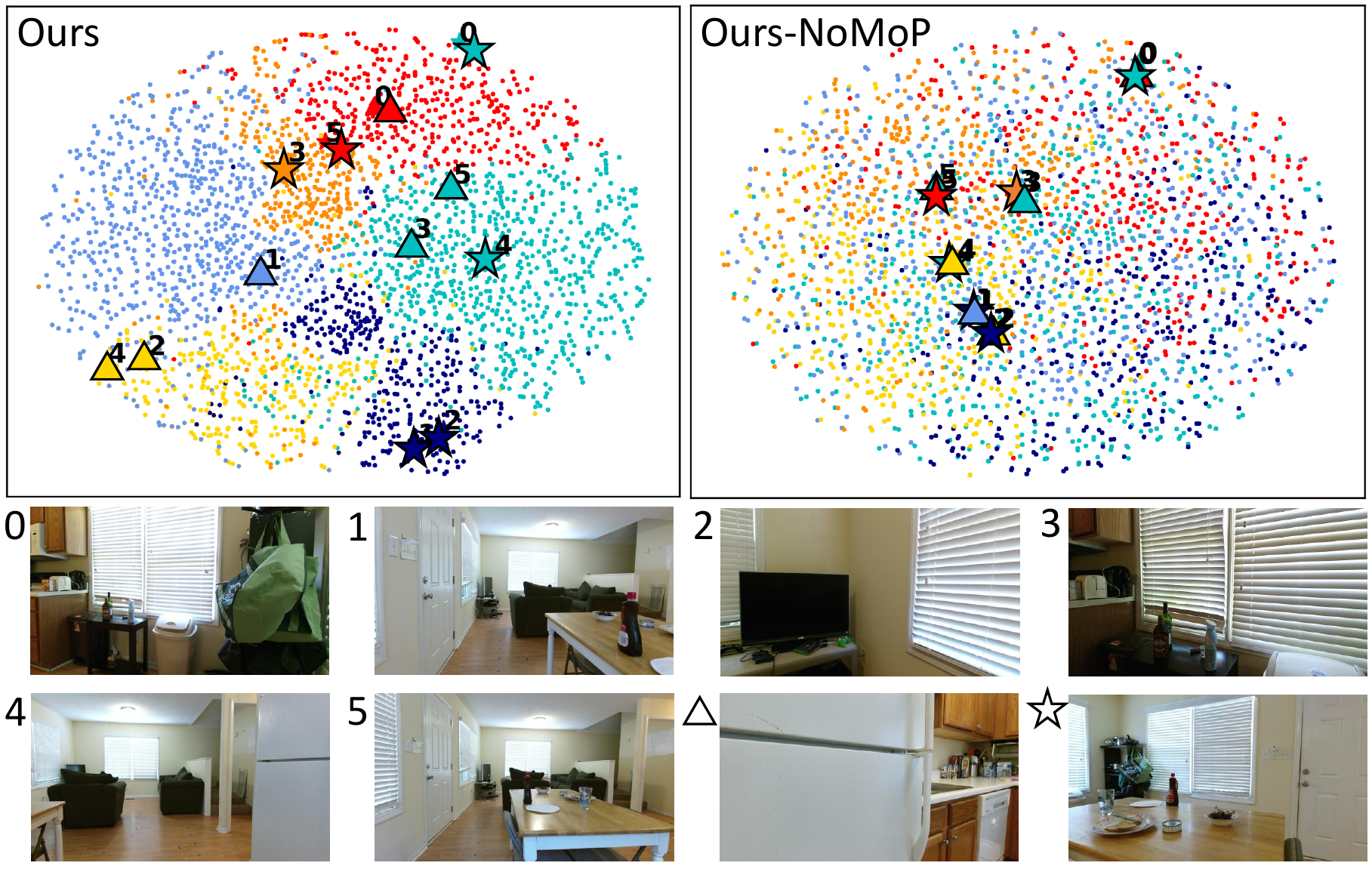}
\end{center}\vspace{-6pt}
   \caption{A t-SNE visualization of the latent space $z\sim q(z|x,g)$ of our model (top-left) and its NoMoP variant (top-right). The color of data points indicates action prediction. Some of the data points are marked with an index of the current (front-view) observation $x$ and a shape symbol indicating the target view $g$ (see the indexing of the corresponding view images at the bottom). The ground-truth action of a data point is visualized as the fill color of the corresponding shape symbol. From the color correspondence between the data points and the co-located shape symbols, our model leads to more accurate action prediction.}
\label{fig:tsne}\vspace{-12pt}
\end{figure}

Table~\ref{tab:semg} reports the breakdown results over various target labels for TD-Semantic with semantic input and our method with RGB and/or depth as input.
We attribute the good performance to the natural design of the learning task in our model.
In TD-Semantic, a deep neural network is learned to predict action cost from
the current observation, the goal and the previous action.
In contrast, our model predicts the next observation from a latent space modeling the dynamics of action-driven observation transition, making it easier to learn an enriched, meaningful representation (see \Figure{tsne}).
Moreover, the variational inference module learns to reason about the surrounding layout based on multi-view observations,
which is helpful for goal-directed decision making even if the goal is represented in an abstract form of a semantic label instead of a view image.

\begin{figure}
\begin{center}
\includegraphics[width=0.95\columnwidth]{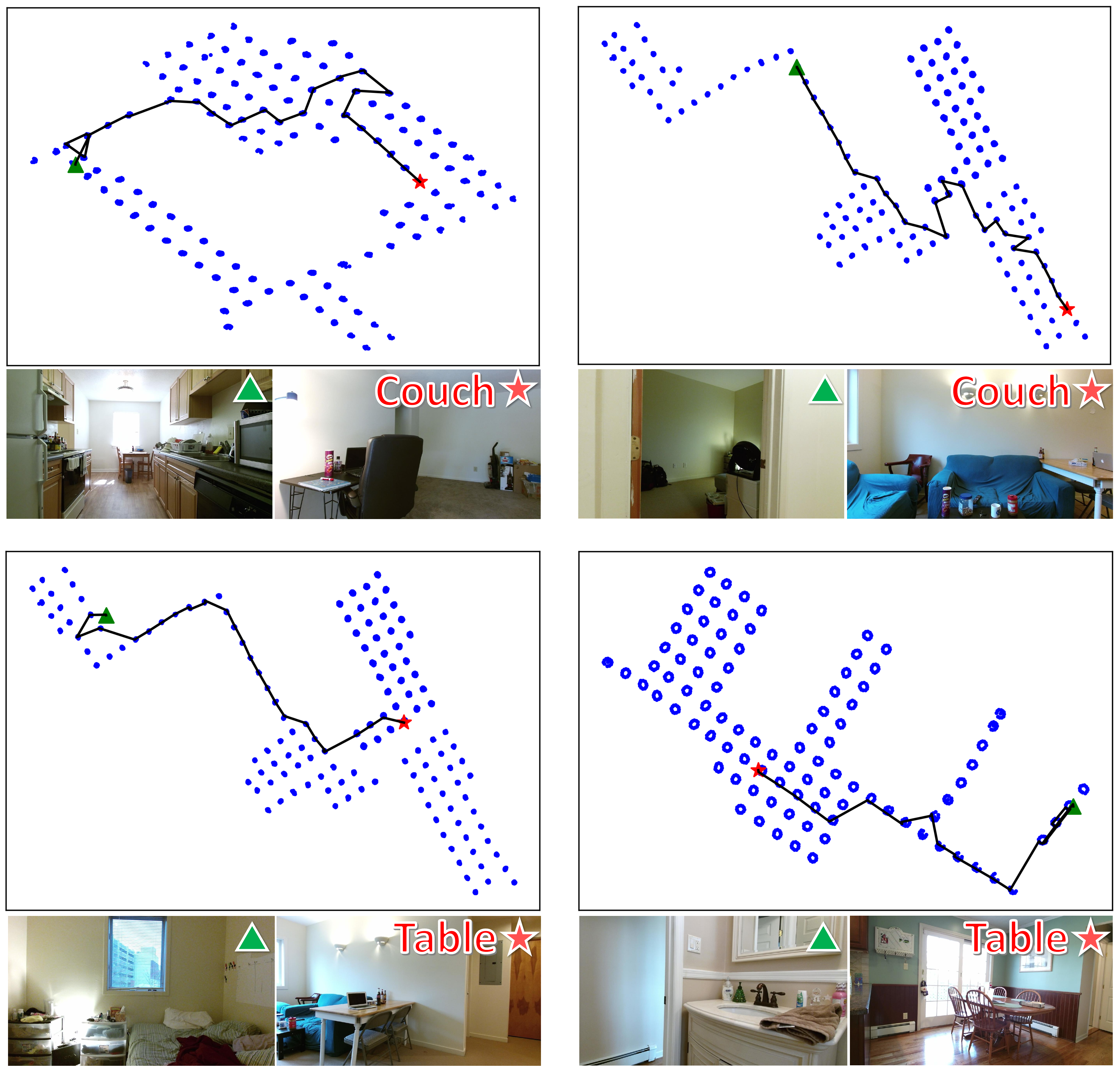}
\end{center}\vspace{-6pt}
   \caption{Visualization of navigation paths on four AVD tasks. Blue dots represent reachable locations in the scene. Green triangles and red stars denote the starting and goal points, respectively.}
\label{fig:path}\vspace{-12pt}
\end{figure}

\vspace{-6pt}
\paragraph{Visualization of the latent space.}
To investigate how well our latent space models the navigation policy
based on the current observation and the target view,
we show in \Figure{tsne} a t-SNE visualization of the latent space $z\sim q(z|x,g)$ learned by our model and its NoMoP baseline.
There are two observations.
First, the latent space of our model exhibits clear structure w.r.t. action predictions (see the color-coding),
making it well suited for navigation decision making. Such expressive latent distribution is facilitated by imposing the mixture-of-posteriors prior conditioned on current observations and next actions $p(z|x,a)$.
In contrast, the latent space constrained by a standard Gaussian prior (NoMoP) $p(z)$ is highly unstructured.
Second, our action prediction is highly accurate (see the correspondence between data point color (action prediction) and shape symbol fill color (ground-truth action))
thanks to the action-driven variational model for NEO estimation and the separate capacity in the network of action prediction.

\vspace{-6pt}
\paragraph{Visualization of navigation paths.}
In \Figure{path}, we visualize the agent paths for four navigation tasks in three unseen scenes from AVD.
Our agent takes close to the shortest paths and achieves successful navigation to the target with no prior knowledge about the environment.
\supl{A visual comparison of navigation paths against alternative methods is provided in the supplemental material.}


\section{Conclusion}
We have presented a generative model for visual navigation that predicts the next action based
on the imagination of the next expected observation.
The NEO generation models the forward dynamics of agent-environment interaction.
The expressive approximation of
the variational posterior as a Mixture of Gaussians leads to a data-efficient model with strong model
generality. We see great potential in incorporating this generative model into a deep RL framework
to address the model imperfection issue in novel scenes.


\section*{Acknowledgments}
We thank Xingyu Xie for fruitful discussions in the early stage of this project.
This work was supported in part by grants from ARO (W911NF-19-1-0069) and NSFC (61772267, 61572507, 61532003, 61622212).

\section{Appendix}
\subsection{Derivation of the variational Bayesian navigation model}
Given the current observation $x$, we opt to generate the next expected observation (NEO) $\hat{x}$ assuming that the next best action $a$ is known \emph{a priori} and is executed. This can be described with a generative model:
\begin{equation}\label{eq:generative}
p_{\theta}(\hat{x}, z|x,a)=p_{\theta}(\hat{x}|z)p_{\theta}(z|x,a)
\end{equation}

We introduce a distribution $q_{\lambda}(z|x,g)$ with parameters $\lambda$ that approximates the true distribution $p_{\theta}(z|x,a)$.
Then we obtain the marginal likelihood of the model:
\begin{equation}\label{eq:lower_bound}
\begin{aligned}
&\log{p_{\theta}(\hat{x}|x,a)}=\log{\int_{z}p_{\theta}(\hat{x},z|x,a)dz}\\
&=\log{\int_{z}p_{\theta}(\hat{x},z|x,a)\frac{q_{\lambda}(z|x,g)}{q_{\lambda}(z|x,g)}dz}\\
&=\log{E_{z\thicksim q_{\lambda}(z|x,g)}[\frac{p_{\theta}(\hat{x},z|x,a)}{q_{\lambda}(z|x,g)}]}\\
&\geq E_{z\thicksim q_{\lambda}(z|x,g)}[\log{\frac{p_{\theta}(\hat{x},z|x,a)}{q_{\lambda}(z|x,g)}}]=\mathcal{L}(\hat{x})
\end{aligned}
\end{equation}
To maximize the marginal likelihood, we maximize its lower bound:
\begin{equation}\label{eq:lower_bound}
\begin{aligned}
&E_{z\thicksim q_{\lambda}(z|x,g)}[\log{\frac{p_{\theta}(\hat{x},z|x,a)}{q_{\lambda}(z|x,g)}}]\\
&=E_{z\thicksim q_{\lambda}(z|x,g)}[\log{\frac{p_{\theta}(\hat{x}|z)p_{\theta}(z|x,a)}{q_{\lambda}(z|x,g)}}]\\
&=E_{z\thicksim q_{\lambda}(z|x,g)}[\log{p_{\theta}(\hat{x}|z)}+\log{\frac{p_{\theta}(z|x,a)}{q_{\lambda}(z|x,g)}}]\\
&=E_{z\thicksim q_{\lambda}(z|x,g)}[\log{p_{\theta}(\hat{x}|z)}]-\mathcal{K}\mathcal{L}[q_{\lambda}(z|x,g)||p_{\theta}(z|a,x)]
\end{aligned}
\end{equation}

This lower bound forms our objective function $-\mathcal{L}(\hat{x})$.
Further, to predict the next best action $a$ based on the generated next expected observation $\hat{x}$,
we train an action classifier $q_{\varphi}(a|x,\hat{x},\tilde{a})$ with $a\thicksim Cat(1/C)$ and $C$ represents the number of action labels. $\tilde{a}$ represents the action being chosen at the previous time step.
We then obtain the following extended objective function:
\begin{equation}\label{eq:obj_func}
\begin{aligned}
\mathcal{J}=&\alpha E_{z\thicksim q_{\lambda}(z|x,g)}[-log{p_{\theta}(\hat{x}|z)}]+\beta \mathcal{K}\mathcal{L}[q_{\lambda}(z|x,g)||p_{\theta}(z|a,x)]\\
&+\gamma E_{a\thicksim p(a)}[-\log{q_{\varphi}(a|x,\hat{x},\tilde{a})}]
\end{aligned}
\end{equation}
where the hyper-parameter $(\alpha, \beta, \gamma)$ tunes the relative importance of the three terms.

\vspace{-6pt}
\subsection{Model architecture and training details}
\begin{figure}[h]
\begin{center}
\includegraphics[width=\linewidth]{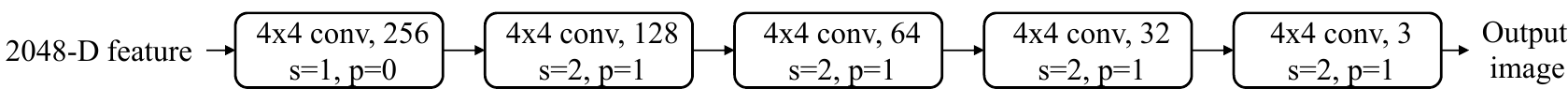}
\end{center}
   \caption{The 5-layer transposed convolutional network for $p_{\theta}(\hat{x}|z)$.
 }\vspace{-5pt}
\label{fig:generator}
\end{figure}

\paragraph{NeoNav} Our navigation model consists of four modules: $q_{\lambda}(z|x,g)$, $p_{\theta}(\hat{x}|z)$, $q_{\varphi}(a|x,\hat{x},\tilde{a})$, and $p_{\theta}(z|x,a)$.
$q_{\lambda}(z|x,g)$ first uses the ResNet-$50$ to extract the features of the current full observation views and the target view, followed by seven fully connected (FC) layers.
The final FC layer connects to two heads to output the mean and variance of a Gaussian distribution from which our latent vector $z$ is sampled.
Given the vector $z$, $p_{\theta}(\hat{x}|z)$ is composed of two FC layers followed by a five-layer transposed convolutional network (see Figure~\ref{fig:generator}).
$q_{\varphi}(a|x,\hat{x},\tilde{a})$ is a four-layer MLP, which takes the
the feature of $x$ from the ResNet-$50$, the feature from $p_{\theta}(\hat{x}|z)$, and the feature extracted from the previous action $\tilde{a}$ by a FC layer as inputs,
and predicts the next action for navigation.
$p_{\theta}(z|x,a)$ is used only in the training stage to regularize the distribution from $q_{\lambda}(z|x,g)$, taking the feature of current observation $x$ (the front-view only) and the ground truth next best action $a^{gt}$ as inputs.

We can either jointly train all submodules within the architecture,
or pretrain the ResNet-$50$ (denoted as $f$) and the $q_{\varphi}(a|x,\hat{x},\tilde{a})$ submodule.
In practice, we found that pre-training leads to faster training of our model.
In this case, let $f(x)$ and $f(\hat{x}^{gt})$ represent the features from the ResNet-$50$,
based on the input of current observation $x$ (front view only) and the ground truth next observation $\hat{x}^{gt}$, respectively.
$q_{\varphi}(a|x,\hat{x},\tilde{a})$ takes $f(x)$, $f(\hat{x}^{gt})$, and the feature extracted from the previous action $\tilde{a}$ by a fully connected layer as inputs.
The loss function for the pre-training is a classification loss $E_{a\thicksim p(a)}[-\log{q_{\varphi}(a|x,\hat{x},\tilde{a})}]$.

We use the SGD optimizer, with a learning rate of $10^{-4}$. We terminate the training when the action prediction accuracy approaches $70\%$. Subsequently, we train the whole model jointly using SGD, with a learning rate of $10^{-5}$.
The motivations of the pre-training are two folds:
1) We hope the ResNet-$50$ extracts more discriminative features for navigation;
2) We hope the $q_{\varphi}(a|x,\hat{x},\tilde{a})$ learns the difference between $x$ and $\hat{x}$ and maps the difference to a driving action from $x$ to $\hat{x}$.
In addition, the architecture in Figure~\ref{fig:generator} can be further simplified.
We can directly use the feature after two FC layers of $p_{\theta}(\hat{x}|z)$,
denoted as $f'(\hat{x})$, and the feature $f(\hat{x}^{gt})$ from ResNet-$50$ to substitute the reconstruction term in our objective function by a $L2$ norm without sacrificing accuracy.
This simplification reduces the number of parameters and hence computational cost.
Our model is trained and tested on a PC with 12 Intel(R) Xeon(R) W-2133 CPU, 3.60 GHz and a Geforce GTX 1080 Ti GPU.

\paragraph{TD-A3C-U, TD-A3C, Gated-LSTM-A3C, I2A} These models are all based on the A3C algorithm~\cite{mnih2016}, which has a discrete policy $\pi(a,x|\vartheta)$ and a value function $v(x|\vartheta)$. A3C optimizes the policy by minimizing the loss function $L_p(\vartheta)=-E_{x_t,a_t,r_t}[\sum_{t=1}^{T}(R_t-v(x_t))\log \pi(a_t,x_t|\vartheta)]-\alpha_h E_{a_t,x_t \sim \pi}[-\log(\pi(a_t,x_t|\vartheta))]$, where the latter term is an entropy regularisation penalty~\cite{jaderberg2016}, which
improves the exploration ability of the model,
and $R_t$ is the discounted accumulative reward defined by $R_t=\sum_{i=0}^{T-t}\tau^i r_{t+i}+v(x_{T+1})$.
The value function is updated by minimizing the loss $L_v(\vartheta)=E_{x_t,r_t}[(R_t-v(x_t))^2]$.

In our setting, TD-A3C, Gated-LSTM-A3C and I2A are trained with strong supervision.
We first pretrain these agents using behavioral cloning (BC, approximate $5e4$ iterations) and then update their policy layers using a shaped reward $r_t=Geo(x_{t-1},g)-Geo(x_t,g)+\zeta$, where $Geo(x_t,g)$ is the geodesic distance between the current observation $x_t$ and the goal $g$ and $\zeta$ is a small constant time penalty~\cite{gordon2019}. In addition, we still keep the navigation action prediction entropy from ground truth. Finally, the overall loss function for these models is $L(\vartheta)=L_p(\vartheta)+\alpha_{v} L_v(\vartheta)+\alpha_{gt} E_{a_t\thicksim p(a)}[-\log(\pi(a_t,x_t|\vartheta))]$, where the hyper-parameters $(\tau, \zeta, \alpha_h, \alpha_v, \alpha_{gt})$ are empirically set as $(0.99, -0.01, 0.01, 0.5, 0.5)$. TD-A3C-U is trained without supervision and the loss function is $L(\vartheta)=L_p(\vartheta)+\alpha_{v} L_v(\vartheta)$, where the related parameters $(\tau, \alpha_h, \alpha_v)$ are empirically set as $(0.99, 0.01, 0.5)$.

During training, we estimate the discounted accumulative rewards and back-propagate through time for every $10$ unrolled time steps with $6$ navigation episodes executed at each time step. Therefore, the batch size is $60$ for each back-propagation.
Each episode terminates when the agent succeeds in finding the target and issues a stop action, or reaches $100$ steps.
A large positive reward $10.0$ will be provided if and only if the agent reaches the goal and the stop action is executed.
In addition, the agent receives a collision penalty $-0.2$ when hitting obstacles.
We use a learning rate $10^{-4}$ and perform $3e5$ training updates for TD-A3C, Gated-LSTM-A3C, I2A and $1e6$ training updates for TD-A3C-U.
We keep the model with the highest training success rate.

\vspace{-6pt}
\subsection{Additional results}
\begin{figure}
\begin{center}
\includegraphics[width=\linewidth]{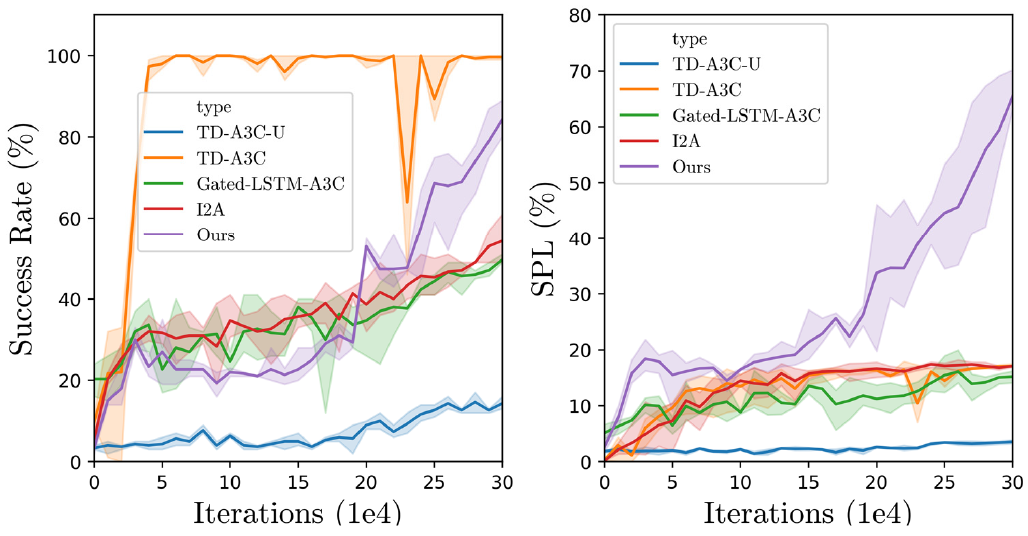}
\end{center}
   \caption{Training curves on AVD depth input. The row shows success rate and SPL, respectively.
 }\vspace{-5pt}
\label{fig:traincurve}
\end{figure}

\paragraph{Training performance on AVD depth input}
We provide the training curves of TD-A3C-U, TD-A3C, Gated-LSTM-A3C, I2A and our model on AVD depth inputs in Figure~\ref{fig:traincurve}. All the models are trained five times with different initializations.
We compute the performance with success rate and SPL every $1e4$ iterations during training.
We use the error band to represent the standard deviation.
TD-A3C, Gated-LSTM-A3C, I2A and our model are all trained with supervision and hence present faster increase than TD-A3C-U in both metrics.
TD-A3C performs better in success rate and worse in SPL than our model,
which results from its high exploration ability.
Our model presents the consistent growth trend in success rate and SPL,
indicating the faster converge to optimal paths than the four A3C-based methods.

\paragraph{Navigation performance on AVD RGB input}
We conduct additional experiments to further compare the performance of all methods, where using RGB images from AVD as inputs.
In Table~\ref{tab:rgb}, all navigation tasks are from the evaluation of cross-scene generalization in the main paper.
Although most models suffer from performance degrading compared to the default depth input, our model with RGB input achieves higher success rate than all other methods.

\begin{table*}
\centering
\caption{Navigation performance (success rate and SPL, in $\%$) comparison on novel scenes from AVD without $stop$ action.}\vspace{-5pt}
\label{tab:rgb}
\scalebox{0.90}{
\begin{tabular}{l|c|c|c|c|c||c}
\cline{1-6}
\hline
\backslashbox{Model}{Target} & Table & Exit & Couch & Refrigerator  & Sink & Avg. \\\cline{1-7}
\hline
\hline
Random Walk & 34.8 / 12.9 &  29.0 / 11.3&  29.8 / 10.8&  27.4 / 10.7&  23.0 / 10.2	&  28.8 / 9.2\\\hline
TD-A3C-U (RGB)&  38.8 / 12.1 &  27.6 / 9.4 &  29.5 / 9.3 &  26.3/ 9.4 &  21.1 / 8.1&
28.7 / 9.7 \\\hline
TD-A3C (RGB)&  40.8 / 3.4 &  29.6 / 2.7 &  27.5 / 3.1 &  21.3/ 2.1 &  27.1 / 1.7&  29.3 / 2.6 \\\hline
Gated-LSTM-A3C (RGB)&  31.0 / 12.9 &  28.0 / 11.3&  23.0 / 8.4&  19.0 / 6.8& 23.0 / 8.1&  24.8 / 9.5\\\cline{1-7}
I2A (RGB)&  42.0 / 16.0 &  27.1 / 7.5&  26.3 / 7.4 &  29.4 / 10.9 & 18.1 / 3.5&  28.6 / 9.1\\\hline
Ours (RGB)& \textbf{ 48.4 / 25.7} &  \textbf{32.4 / 16.7} &  \textbf{28.7 / 16.8} &  \textbf{37.9 / 18.8} &   \textbf{38.2 / 26.4} &  \textbf{37.1 / 20.9} \\\hline
\end{tabular}}
\end{table*}

\begin{table*}
\centering
\caption{Navigation performance (success rate and SPL, in $\%$) comparison on novel scenes from AVD without $stop$ action.}\vspace{-5pt}
\label{tab:reverse}
\scalebox{0.90}{
\begin{tabular}{l|c|c|c|c|c||c}
\cline{1-6}
\hline
\backslashbox{Model}{Target} & Table & Exit & Couch & Refrigerator  & Sink & Avg. \\\cline{1-7}
\hline
\hline
Random Walk & 26.2 / 8.1 &  26.6 / 10.8&  28.1 / 11.7&  32.9 / 11.3&  25.1 / 8.0	&  27.8 / 10.0\\\hline
TD-A3C-U&  26.7 / 7.2 &  28.3 / 7.9&  30.1 / 9.9 &  28.4 / 10.1 &  22.9 / 9.7&  27.3 / 9.0 \\\hline
TD-A3C&  25.3 / 3.9 &  29.6 / 7.0 &  24.1 / 0.8 &  29.3 / 3.8 &  24.4 / 2.3&  26.5 / 3.6\\\hline
Gated-LSTM-A3C&  15.2/ 7.6 &  13.5 / 4.3&  8.6/ 3.0&  14.2 / 6.6& 8.3 / 3.0&  12.0 / 4.9\\\cline{1-7}
I2A&  \textbf{37.9} / 12.7 &  31.8 / 10.4&  29.9 / 8.4 &  29.3 / 8.6 & 23.1 / 11.1&  30.4 / 10.2\\\hline
Ours&31.3/\textbf{14.9} & \textbf{37.3 / 20.3}& \textbf{33.4 / 17.0}& \textbf{23.2 / 12.3}& \textbf{36.8 /17.2} &  \textbf{32.4 / 16.3}\\\hline
\end{tabular}}
\end{table*}

\begin{figure*}
\begin{center}
\includegraphics[width=.95\linewidth]{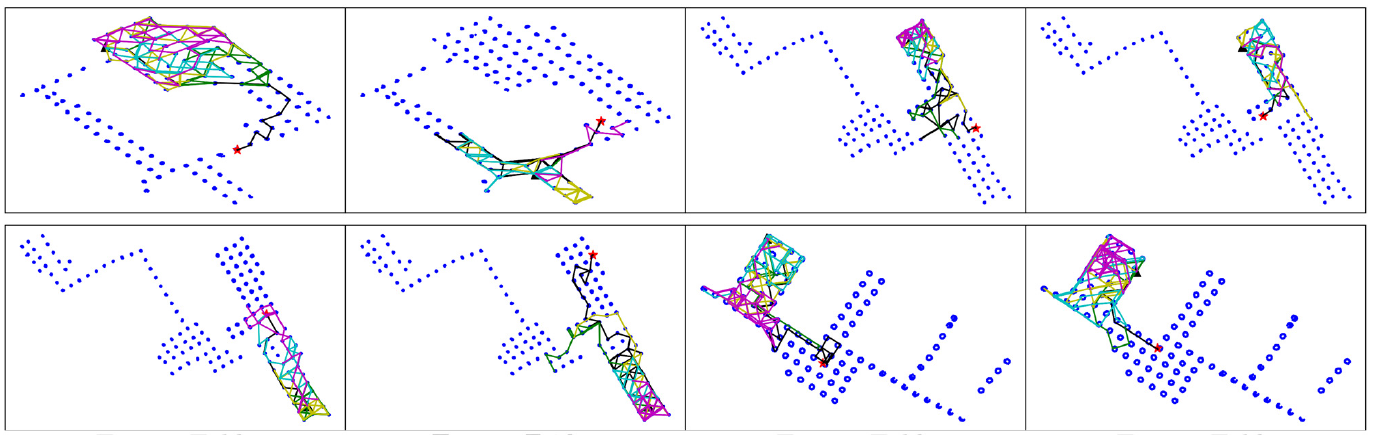}
\end{center}\vspace{-5pt}
   \caption{Visual comparison of navigation paths among our method, TD-A3C-U, TD-A3C, Gated-LSTM-A3C, and I2A, over eight different navigation tasks. Blue dots represent the reachable positions in the scenes. Black triangles and red stars denote starting and goal points, respectively. TD-A3C-U, TD-A3C, Gated-LSTM-A3C, and I2A choose the magenta, the green, the cyan and the yellow paths, respectively. Our agent takes the black paths and is able to successfully navigate to the goals.
 }\vspace{-12pt}
\label{fig:cmypath}
\end{figure*}

\begin{figure*}
\begin{center}
\includegraphics[width=.95\linewidth]{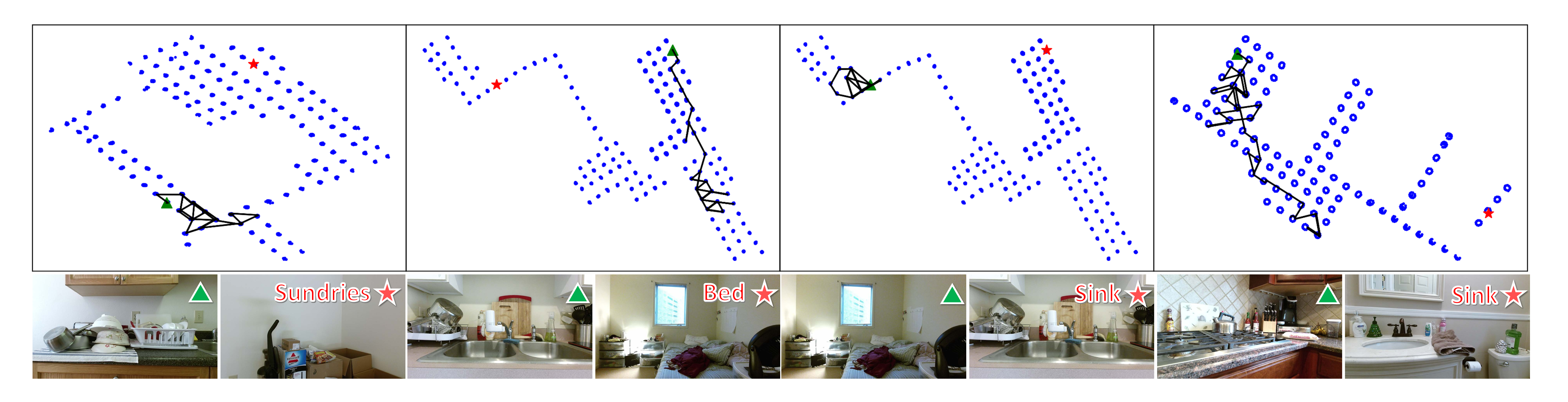}
\end{center}\vspace{-5pt}
   \caption{Visualization of some typical failure cases of our method in four navigation tasks from AVD. The blue dots represent reachable locations in the scene. Green triangles and red stars denote starting and goal points, respectively.
 }\vspace{-12pt}
\label{fig:failure}
\end{figure*}

\paragraph{Path-reversing generalization}
This experiment is a supplement to the evaluation of cross-scene generalization on AVD in the main paper.
Here, we reverse the start and target views of each navigation task.
The results are reported in Table~\ref{tab:reverse}.
The scenario in which both scenes and target objects are novel is quite
challenging, and all models perform worse for these path reversing tasks than the normal cases.
Therefore, although depth information can provide strong cues for room layout,
such input tends to be too strong for learning a more general model.

\begin{table}
\centering
\caption{Comparing average success rate (in $\%$) of our model and TD-Semantic for navigation driven by semantic labels, based on various input modalities from AVD.}\vspace{-5pt}
\label{tab:TE1}
\scalebox{0.90}{\begin{tabular}{c|c|c|c|c|c|c}
\hline
\cline{1-7}
\multicolumn{3}{ c| }{Ours}&\multicolumn{4}{ |c }{TD-Semantic} \\\cline{1-7}
 RGB &Depth &RGBD &RGB &Depth &RGBD&Det.\\\cline{1-7}
\hline
54 & 66 & 60 & 31 & 31 & 28 &48 \\\cline{1-7}
\hline
\end{tabular}}\vspace{-12pt}
\end{table}

\paragraph{Navigation driven by semantic labels}
We report the average performances of our method and TD-Semantic for semantic-driven navigation tasks.
Both models are trained on AVD and tested on three unseen scenes.
Our results are based on $2000$ navigation tasks sampled from the test scenes and the success criterion is within $5$ steps to the goal as in~\cite{mousavian2018}.
As shown in Table~\ref{tab:TE1},
our method outperforms the TD-Semantic for all input modalities.

\paragraph{Visual comparison of navigation paths}
We visualize the agent trajectories by our model and four learning-based alternatives (TD-A3C-U, TD-A3C, Gated-LSTM-A3C, and I2A), for eight different navigation tasks (see Figure~\ref{fig:cmypath}).
These are all relatively challenging tasks in which the agent starts from a location from where the desired goal is completely invisible to the agent.
For all the eight tasks, most alternative methods fail to reach the goals. In contrast, our agent is able to navigate to the goals successfully.

\paragraph{Visualization of failure cases}
We visualize the trajectories for some failure cases of our method (see Figure~\ref{fig:failure}).
These tasks are all characterized by unknown scenes, novel targets and far distances between the start points and the goals.
Our agent fails to finish these navigation tasks within the maximum number
of steps $(100)$.
The problems are navigating around
tight spaces (e.g, the cramped kitchen where the first trajectory starts),
getting stuck in the corner (see the third trajectory),
and thrashing around in space without making progress (see the second and fourth trajectories).

\small
\bibliography{navigation}
\bibliographystyle{aaai}

\end{document}